\documentclass[twoside]{article}
\usepackage{amsfonts,amssymb,amsbsy,textcomp,marvosym,picins,amsmath,caption,threeparttable,amsthm,subfigure}
\usepackage{eurosym,mathrsfs,fancyhdr,CJK,multicol,graphics,indentfirst,color,bm,upgreek,booktabs,graphicx}
\usepackage[noend]{algorithm}
\usepackage[noend]{algorithmic}
\def\footnoterule{\kern 1mm \hrule width 10cm \kern 2mm}

\allowdisplaybreaks
\catcode`@=11
\def\title#1{\vspace{3mm}\begin{flushleft}\vglue-.1cm\Large\bf\boldmath\protect\baselineskip=18pt plus.2pt minus.1pt #1
\end{flushleft}\vspace{1mm} }
\def\author#1{\begin{flushleft}\normalsize #1\end{flushleft}\vspace*{-4pt} \vspace{3mm}}
\def\address#1#2{\begin{flushleft}\vglue-.35cm${}^{#1}$\small\it #2\vglue-.35cm\end{flushleft}\vspace{-2mm}\par}

\catcode`@=11
\def\section{\@startsection{section}{1}{\z@}%
 {-3ex \@plus -.3ex \@minus -.2ex}%
 {2.2ex \@plus.2ex}%
{\normalfont\normalsize\protect\baselineskip=14.5pt plus.2pt minus.2pt\bfseries}}
\def\subsection{\@startsection{subsection}{2}{\z@}%
 {-3ex\@plus -.2ex \@minus -.2ex}%
 {2ex \@plus.2ex}%
{\normalfont\normalsize\protect\baselineskip=12.5pt plus.2pt minus.2pt\bfseries}}
\def\subsubsection{\@startsection{subsubsection}{3}{\z@}%
 {-2.2ex\@plus -.21ex \@minus -.2ex}%
 {1.4ex \@plus.2ex}
{\normalfont\normalsize\protect\baselineskip=12pt plus.2pt minus.2pt\sl}}

\begin{document}
\begin{CJK*}{GBK}{song}
\thispagestyle{empty}
\vspace*{-13mm}
\noindent {\small }
%===========================================================
\vspace*{2mm}

\title{PVSS: A Progressive Vehicle Search System for Video Surveillance Networks}
\author{Xin-Chen Liu$^{1}$, Wu Liu$^{1}$, Hua-Dong Ma$^{2}$, and Shuang-Qun Li$^{2}$}

\address{1}{JD AI Research, JD.com, Beijing 100101, China}
\address{2}{Beijing Key Laboratory of Intelligent Telecommunication Software and Multimedia, Beijing University of Posts and Telecommunications, Beijing 100876, China}
\footnotetext{{}\\[-4mm]\indent\quad preprint for arXiv}

\noindent {\small\bf Abstract} \quad  {\small \textcolor{blue}{
This paper is focused on the task of searching for a specific vehicle that appeared in the surveillance networks. Existing methods usually assume the vehicle images are well cropped from the surveillance videos, then use visual attributes, like colors and types, or license plate numbers to match the target vehicle in the image set. However, a complete vehicle search system should consider the problems of vehicle detection, representation, indexing, storage, matching, and so on. Besides, attribute-based search cannot accurately find the same vehicle due to intra-instance changes in different cameras and the extremely uncertain environment. Moreover, the license plates may be misrecognized in surveillance scenes due to the low resolution and noise. In this paper, a Progressive Vehicle Search System, named as PVSS, is designed to solve the above problems. PVSS is constituted of three modules: the crawler,  the indexer, and the searcher. The vehicle crawler aims to detect and track vehicles in surveillance videos and transfer the captured vehicle images, metadata and contextual information to the server or cloud. Then multi-grained attributes, such as the visual features and license plate fingerprints, are extracted and indexed by the vehicle indexer. At last, a query triplet with an input vehicle image, the time range, and the spatial scope is taken as the input by the vehicle searcher.
The target vehicle will be searched in the database by a progressive process. Extensive experiments on the public dataset from a real surveillance network validate the effectiveness of the PVSS.
}}

\vspace*{3mm}


\vspace*{4mm}

\end{CJK*}
\baselineskip=18pt plus.2pt minus.2pt
\parskip=0pt plus.2pt minus0.2pt
\begin{multicols}{2}

\section{Introduction}
\begin{figure*}[t]
  \centering
  {\includegraphics[width=0.8\textwidth]{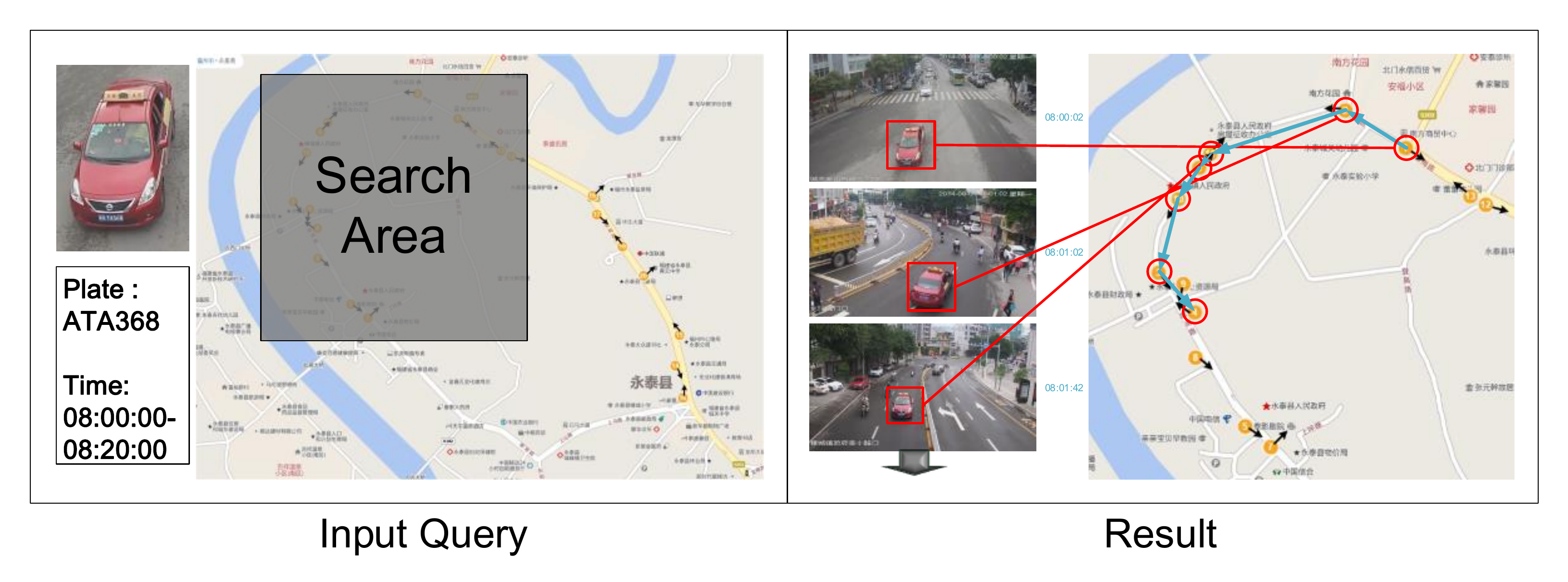}}
  \caption{A typical example of vehicle search. Given a specific vehicle, a time interval, and the spatial scope, the system returns where and when the vehicle appeared in the surveillance network.}
  \label{fig:figure1}
\end{figure*}

%Background
\label{sec:intro}
Physical object search, which aims to find an object sensed by ubiquitous sensor networks like surveillance networks, is one of the most important services provided by the Internet of Things (IoT)~\cite{ma2017progressive}.
Vehicle, including car, bus, truck, etc., is one type of the most common objects in video surveillance networks.
So vehicle search system has many potential applications in the era of IoT.
The search engines of the Internet, e.g., Google, YouTuBe, and Amazon's search engine, can assist us in looking for webpages, images, videos, and products in the information space or cyber space ~\cite{brin1998anatomy, gan2016you}, while the task of vehicle search engine is to find the target vehicle in the physical space.
Vehicle search system can provide pervasive applications such as intelligent transportation~\cite{zhang2011data} and automatic surveillance~\cite{valera2005intelligent}.
Fig.~\ref{fig:figure1} shows an example, in which the user can input a query vehicle, search area and time interval, the system can return the locations and timestamps of the target.

%Figure 1
%\begin{figure}[t]
%  \centering
%  {\includegraphics[width=\columnwidth]{figure1}}
%  \caption{A typical example of vehicle search. Given a specific vehicle, a time interval, and the spatial scope, the system returns where and when the vehicle appeared in the surveillance network.}
%  \label{fig:figure1}
%\end{figure}

%Existing methods and challenges
Early vehicle retrieval methods and systems are mainly focused on the attribute-based framework~\cite{feris2011attribute, feris2012vehicle, gan2016learning}.
They first classify vehicles by types, models, and colors, then index and retrieve them with the assigned attributes.
Recently, vehicle search research is focused on content-based vehicle matching, also known as vehicle Re-Identification (Re-Id), which uses the content of images to fine vehicles in the database~\cite{liu2016vehiclereid, liu2016DRDL}.
Besides, multi-modal contextual information like spatiotemporal information is also explored to assist vehicle Re-Id~\cite{liu2016provid, shen2017learning, Wang_2017_ICCV, gan2017vqs}.
With the development of representation models, such as hand-crafted descriptors and Convolutional Neural Network (CNN), these methods obtain significant improvement.
However, it is difficult to precisely find the specific vehicle only based on attributes because of the intra-instance changes in different cameras and the minor inter-instance difference between similar vehicles.
Furthermore, existing vehicle Re-Id approaches assume that the vehicle images have been well cropped and aligned from the video frames.
Therefore, they only consider the feature extraction and one-to-N matching for the vehicle images.
Nevertheless, a vehicle search engine, as a complex system, must consist of many components like vehicle extraction, representation, indexing, and retrieval.
Moreover, both the accuracy and efficiency should be considered when designing the system.

%Figure 2
\begin{figure*}[t]
  \centering
  {\includegraphics[width=0.95\textwidth]{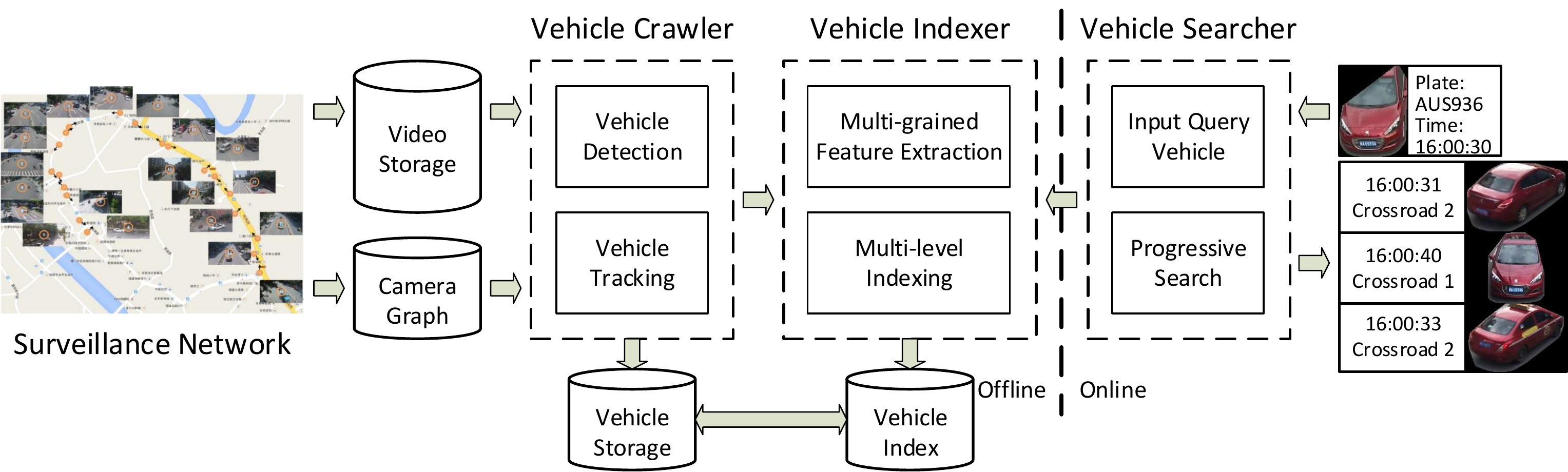}}
  \caption{The architecture of the progressive vehicle search system.}
  \label{fig:figure2}
\end{figure*}

Towards this end, we design a progressive vehicle search system, named as PVSS, in this paper.
PVSS contains three key modules: the crawler of vehicle data, the vehicle indexer based on multi-grained features, and the progressive vehicle searcher.
To guarantee high accuracy and efficiency during search, a series of data structures are designed for the vehicle search system.
In the crawler, not only visual contents but also contextual information are extracted from the surveillance networks.
Then the multimodal data is exploited by deep learning based models to obtain discriminative and robust features of vehicles, which is then organized by the multi-level indexes.
In the search process, the vehicle is searched in a progressive manner, including the from-coarse-to-fine search in the feature domain and the from-near-to-distant search in the physical space.
At last, extensive experiments on a large-scale vehicle search dataset collected from real-world surveillance network shows the state-of-the-art results of the proposed system.

Compared with our previous conference paper~\cite{liu2018vehicle}, we provide more analysis on contextual information such as the spatiotemporal information in surveillance networks.
For example, we discuss the temporal distance between neighboring cameras in the surveillance network by analyzing the travel time of vehicles in our collected data.
We also compare the characteristics of vehicles to that of persons which have been studied in related work.
Based on the analysis of spatiotemporal information of vehicles in surveillance networks, we propose a new camera neighboring graph compared to~\cite{liu2018vehicle}.
Particularly, in~\cite{liu2018vehicle} we only adopted the fixed spatial distance between neighboring cameras as the weights of edges in the graph, which is too simple to model the spatiotemporal cues.
In this new manuscript, we also use the temporal distance between neighboring cameras learned from training data to modeling the spatiotemporal relations, which further improve the performance of the system.

%Related Works
\section{Related Work}
\label{sec:related}
\subsection{Multimedia Retreival}
In the past two decades, content based multimedia retrieval (CBMR) has been extensively studied~\cite{sivic2003video, lew2006content, hu2011survey, mei2014multimedia, liu2014instant, ZhengYT18, gan2016you, gan2016webly}.
CBMR methods usually extract visual features from images or videos and estimate the similarity between the query and source in the database.
For examples, Video Google was proposed by Sivic and Zisserman to achieve object search in videos with the idea of webpage retrieval ~\cite{sivic2003video}.
Lin \textit{et al.}~\cite{lin2013investigating} exploited the 3-D representation models for content based vehicle search.
Farhadi \textit{et al.}~\cite{farhadi2009describing} proposed to represent the appearance of objects by their attributes for image retrieval.
Zheng \textit{et al.}~\cite{ZhengWLT15} proposed a large-scale image retrieval method with an effective visual model and efficient index structures.
Liu \textit{et al.}~\cite{liu2014instant} designed an instant video search system for movies search on mobile devices.
However, different from existing CBMR task, only depending on visual features, i.e. the appearance of vehicles, cannot give precise results because of the minor inter-class differences between very similar vehicles and varied intra-instance changes in different cameras.

\subsection{Person Re-Id and Search}
Content based person Re-Id has been studied for several years~\cite{farenzena2010person, gong2014person, ZhengYH16}.
Existing person Re-Id approaches usually assume the persons have been detected and extracted from the video frames.
The main topics include visual features learning from images and discriminative metrics for feature embedding~\cite{zheng2015scalable}.
Besides person Re-Id, attributes and context information are also used for person retrieval.
For examples, Feris \textit{et al.}~\cite{feris2014attribute} proposed a system for attribute-based people search in surveillance environments.
Xu \textit{et al.}~\cite{xu2013graph} designed an object browse and retrieval system, which integrated vision features and spatial-temporal cues by a graph model for retrieval of pedestrians and cyclists.

\subsection{Vehicle Re-Id and Search}
In recent years, vehicle search is mainly focused on content based vehicle Re-Id, which aims to find the target vehicle from the database with a query vehicle image~\cite{liu2016DRDL, liu2016vehiclereid}.
For example, Liu \textit{et al.}~\cite{liu2016DRDL} proposed a deep CNN based method, named Deep Relative Distance Learning, to jointly learn visual features and metric mapping for vehicle Re-Id.
Besides appearance features, the contextual information such as license plates and spatiotemporal records is also used for vehicle Re-Id.
For examples, Liu \textit{et al.}~\cite{liu2016provid} proposed a progressive vehicle search method which exploits image features, license plates, and contextual information in a progressive manner.
Wang \textit{et al.}~\cite{Wang_2017_ICCV} proposed a framework to learn local landmarks and global features of vehicles and refine the results with a spatiotemporal regularization model.
Similar to person Re-Id, existing vehicle Re-Id methods also assume that the vehicle images have been detected and well aligned from video frames.
Therefore, they only consider the feature representation and similarity metrics for image matching.
However, to build a complete search system, we consider not only the problems for content based vehicle Re-Id but also the tasks of data acquisition, organization, and retrieval.

%Table 1
\begin{table*}[t]
\caption{Vehicle Track Metadata.}
\label{tab:table1}
\centering
\begin{tabular}{p{80pt}p{25pt}p{300pt}} \hline
\textbf{Name}       	&   \textbf{Type}   	&   \textbf{Description}    \\  \hline
Camera ID           	&   int             		&   The unique ID of the camera that captures the track.   \\  \hline
Vehicle ID           	&   long            		&   The unique ID of the vehicle track.  \\  \hline
Frame ID            	&   long            		&   The ID of the first frame in the vehicle track.  \\  \hline
Track Length        	&   int             		&   The frame count of the vehicle track. \\  \hline
Trajectory          	&   point[]         		&   The point sequence of the vehicle track.  \\  \hline
Visual Features     	&   float[]         		&   The multi-grained visual features extracted from the vehicle track.    \\  \hline
Duration               	&   float           		&   The time duration of the vehicle track.        \\  \hline
Plate               	&   string          		&   The license plate string of the object (if recognized).      \\  \hline
\end{tabular}
\end{table*}

%Section 2
\section{Overview}
\label{sec:arch}
Fig.~\ref{fig:figure2} illustrates the overall architecture of the PVSS system.
It contains three moduels:
\begin{itemize}
  \item \textbf{The offline vehicle crawler} receives the video streams from surveillance cameras and crops vehicle image sequences from video frames.
  \item \textbf{The vehicle indexer} extracts multi-grained visual features from vehicle tracks and constructs the multi-level indexes for efficient search
  \item \textbf{The online vehicle searcher} performs the progressive search process with the multi-level indexes in both the feature domain and the spatiotemporal space.
\end{itemize}
Before introducing the details of each component, we first present the main data structures of PVSS in next section.

%Section 2.1
\section{Data Structures}
\label{subsec:data}
The data that we can utilize is diverse and in multiple modalities.
Various semantic contents like vehicle plates, types, colors, and visual features can be extracted in online or offline manner as in \cite{yang2015compcars, liu2016provid}.
The data modalities include text, digits, coordinates, structures, and so on.
The topology and spatiotemporal context of surveillance networks can be more complex data structures such as graphs.
Therefore, these data should be described in proper structures, which are effective for retrieval and flexible for extension.
In this section, we first introduce the vehicle track metadata, which is to describe the image sequences of vehicles captured by surveillance cameras.
Then, the camera table is designed to index the vehicle track metadata for each camera.
At last, we build a camera neighboring graph to represent the spatial topology of the surveillance networks.

%Section 2.1.1
\subsection{Vehicle Track Metadata}
\label{subsubsec:metadata}

According to the variety of video contents and extraction approach, the vehicle track metadata is proposed to describe vehicle image sequences which are obtained from cameras.
Table 1 lists the attributes and descriptions of the metadata in detail.
In our system, the vehicle tracks are extracted by the vehicle crawler frame by frame, which will be presented in Section~\ref{subsec:crawler}.
The object tracking method is used to group the images of the same vehicle in neighbor frames as an instance of vehicle track.
As in Table 1, the unique Camera ID and Vehicle ID specify an unique vehicle.
Among these attributes, the visual features are the most important information to represent the multi-grained visual representation of each vehicle, which are utilized in the indexing and search procedures.
The extraction of visual features will be given in Section~\ref{subsec:index}.

%Section 2.1.2
\subsection{Camera Table}
\label{subsubsec:camtable}
After the generation of vehicle track metadata, the storing and indexing of these data should be considered.
In out system, the camera table is designed to index instances of vehicle track metadata for each camera.

For each camera, we allocate a camera table to index the vehicle track metadata extracted from this camera.
The videos are processed by the order of time, so the metadata instances are also generated by the order of time and appended to the tails of camera tables.
This keeps the entries of camera tables in the ascending order.
Fig. 3 shows the structure of the camera table.
In the real implementation, the camera table can be implemented by relational databases like MySQL or distributed databases like HBase in the data center.
When the scale of camera tables grows up, the tables will be organized in a tree-like structure for efficient index and search.

%Figure 3
%\begin{figure}[ht]
%  \centering
%  {\includegraphics[width=0.9\columnwidth]{figure3}}
%  \caption{The structure of the camera table.}
%  \label{fig:figure3}
%\end{figure}

\vspace{2mm}

\begin{center}
\includegraphics[width=0.9\columnwidth]{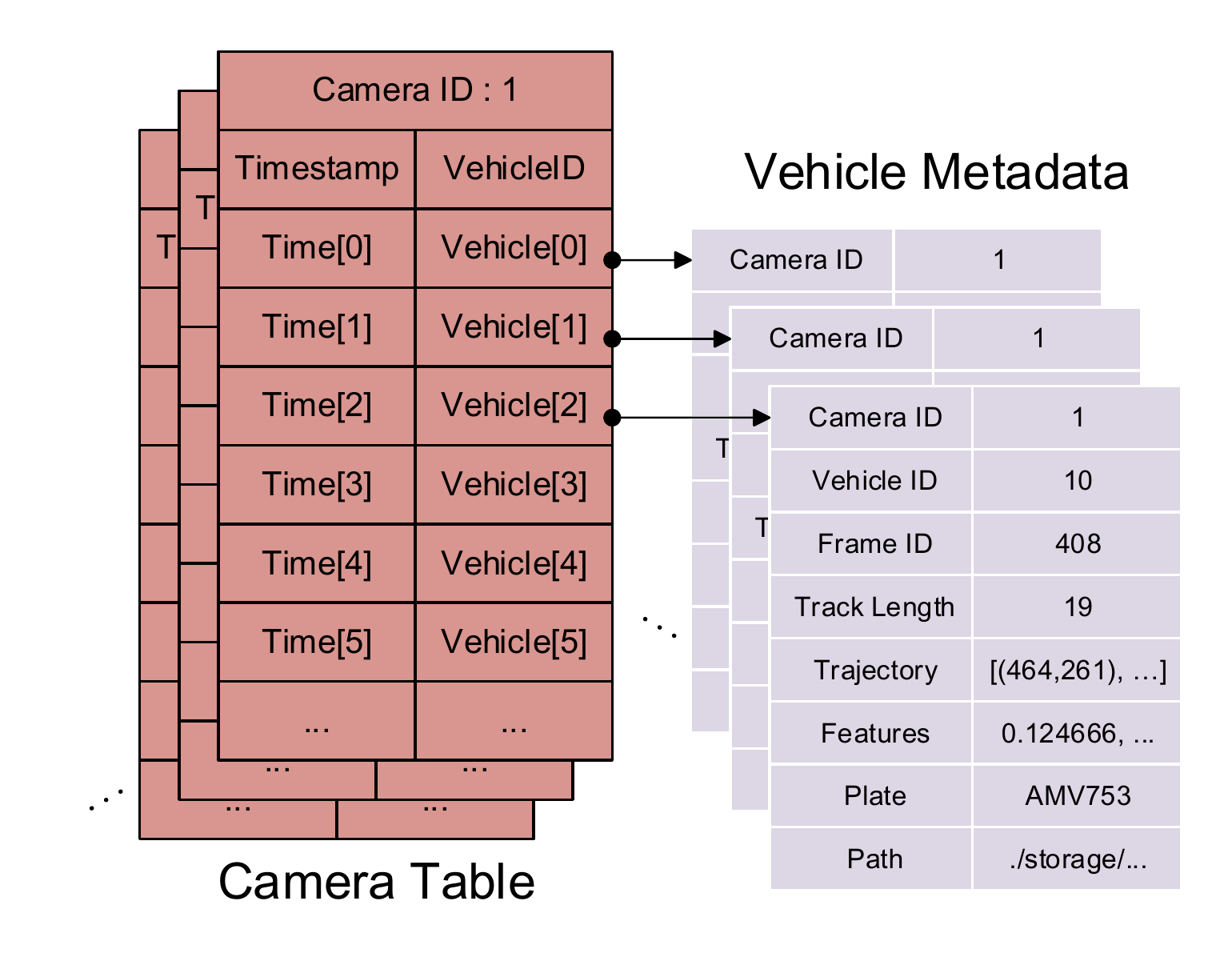}\\
\vspace{2mm}
\parbox[c]{8.3cm}{\footnotesize{Fig.3.~}  The structure of the camera table. }%\vspace*{.2mm}
\label{fig:figure3}
\end{center}

\vspace{1mm}

%Section 2.1.4
\subsection{Camera Neighboring Graph}
\label{subsubsec:graph}

\subsubsection{Topology Construction}
The camera neighboring graph records the geo-locations of cameras and the topology of the surveillance networks, which is obtained from the infrastructure companies and the map services.

We define the graph as a directed graph $G = <N, E, W>$.
The graph is composed by the node set $N = \{n_1, ..., n_C\}$, the edge set $E = \{e_{i,j}\}$, and the weight set $W = \{w_{i,j}\}$.
Fig.~\ref{fig:figure4} illustrates an example of the camera neighboring graph which is built from a subset of real-world surveillance network.
The nodes represent the set of cameras, which consist of the GPS coordinates and settings of cameras.
The edges are the set of directed connections between neighboring cameras.
The edges are determined not only by the topology of the city roads but also by the heading directions and fields of view (FOV) of cameras.
So we define the view-connected edge as below:

{\bf Definition 1} (View-connected edge). A view-connected edge $e_{i,j} = ( n_i, n_j )$ connects a pair of cameras in $N$, if an vehicle can reappear in the FOV of camera $j$ directly after appearing in the FOV of camera $i$, then there is a view-connected edge $e_{i,j}$ from $n_i$ to $n_j$.

%Algorithm 1 gives the construction procedure of $M$.
%
%%Algorithm 1
%\begin{center}
%\parbox[c]{8.3cm}{\footnotesize{\textbf{Algorithm 1~}} Adjacent Matrix Construction.}%\vspace*{.2mm}
%    \begin{algorithmic}[1]
%        \STATE \textbf{Input:} travel distance matrix $M_d = \{d_{i,j}\}$, labeled direction number set $Dn = \{dn_i\}$
%        \STATE \textbf{Output:} adjacent matrix $M_e = \{e_{i,j}\}$
%        \STATE \textbf{Initialize:} $ M_e = \{e_{i,j} = 0\}$
%        \FOR {$i = 1 \rightarrow n$}
%            \FOR {$k = 1 \rightarrow dn_i$}
%                \STATE $min = MAX\_VALUE$
%                \STATE $index = -1$
%                \FOR {$j = 1 \rightarrow n$ }
%                    \IF {$ min < d_{i,j} $}
%                        \STATE $min = d_{i,j}$
%                        \STATE $index = j$
%                        \STATE $d_{i,j} = MAX\_VALUE$
%                    \ENDIF
%                \ENDFOR
%                \STATE $e_{i,index} = 1$
%            \ENDFOR
%        \ENDFOR
%        \RETURN $M_e$
%    \end{algorithmic}
%\end{center}

%Figure 4
\setcounter{figure}{3}
\begin{figure*}[t]
  \centering
  {\includegraphics[width=0.8\textwidth]{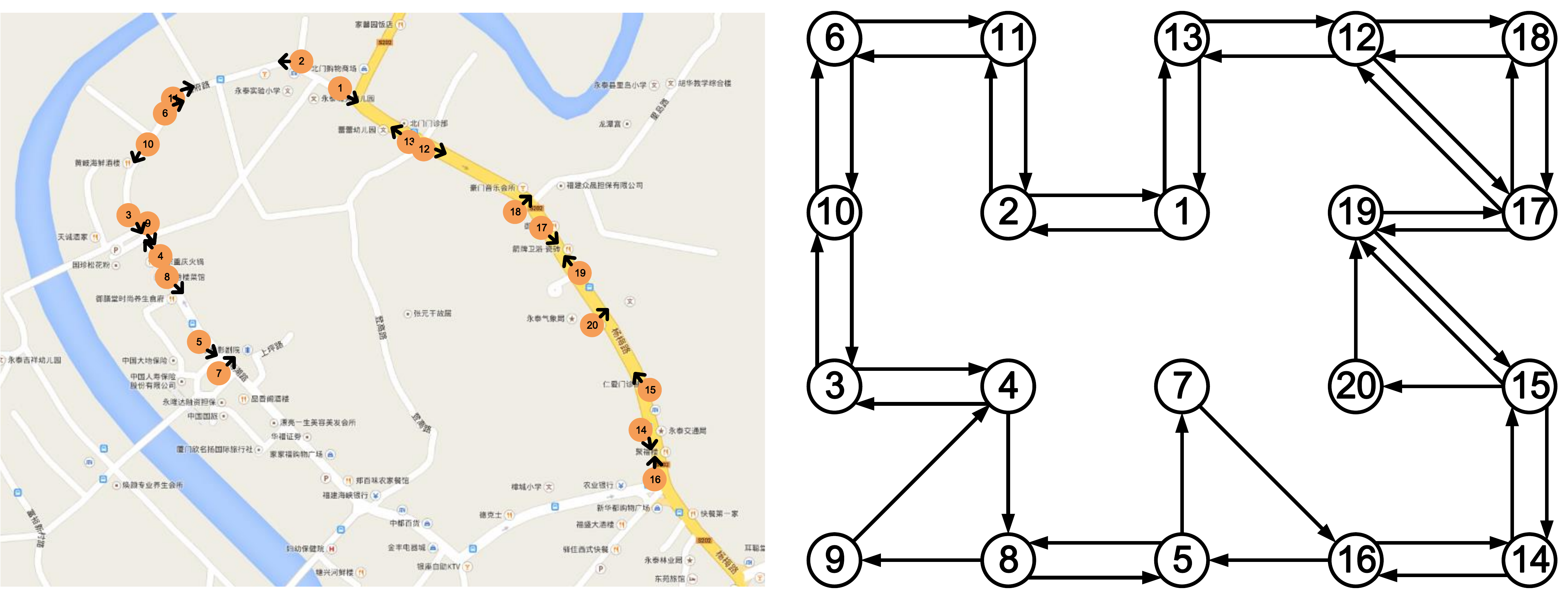}}
  \caption{An example of the camera neighboring graph. The left image is the camera locations and the city map of a real-world surveillance network. The right is the graph abstracted from the network.}
  \label{fig:figure4}
\end{figure*}

\subsubsection{Weight Modeling}
The weight set $W$ of $G$ contains two parts.
The first part is $W_t$.
It stores the spatial distances of neighboring cameras, which can be obtained from map services like Google Map.
The second part is $W_s$ which contains the temporal distances between neighboring cameras learned from training data.
Here we will give details about the learning of $W_s$.

Several works have proposed models to estimate the travel time in surveillance networks.
The author of~\cite{liu2014vehicle} proposed a graph-based vehicle search model.
According to this model, the weight of an edge is modeled by the mean time cost of all vehicles that traveled the edge during the search time.
When given a search time interval, the history records in the time interval are used to compute the mean time cost in this time interval.
Xu \textit{et al.}~\cite{xu2013graph} proposed a graph model for related object search in a campus.
This model estimates the time delay between cameras using object reappearance.
It is assumed that the speed of an object changes slightly, so the time delay is negatively linearly correlative to the travel speed.
Using the labeled data collected from the surveillance network, a linear model of time delay and optical flow is learned with a standard regression method.

\begin{figure*}[t]
  \centering
  {\includegraphics[width=\textwidth]{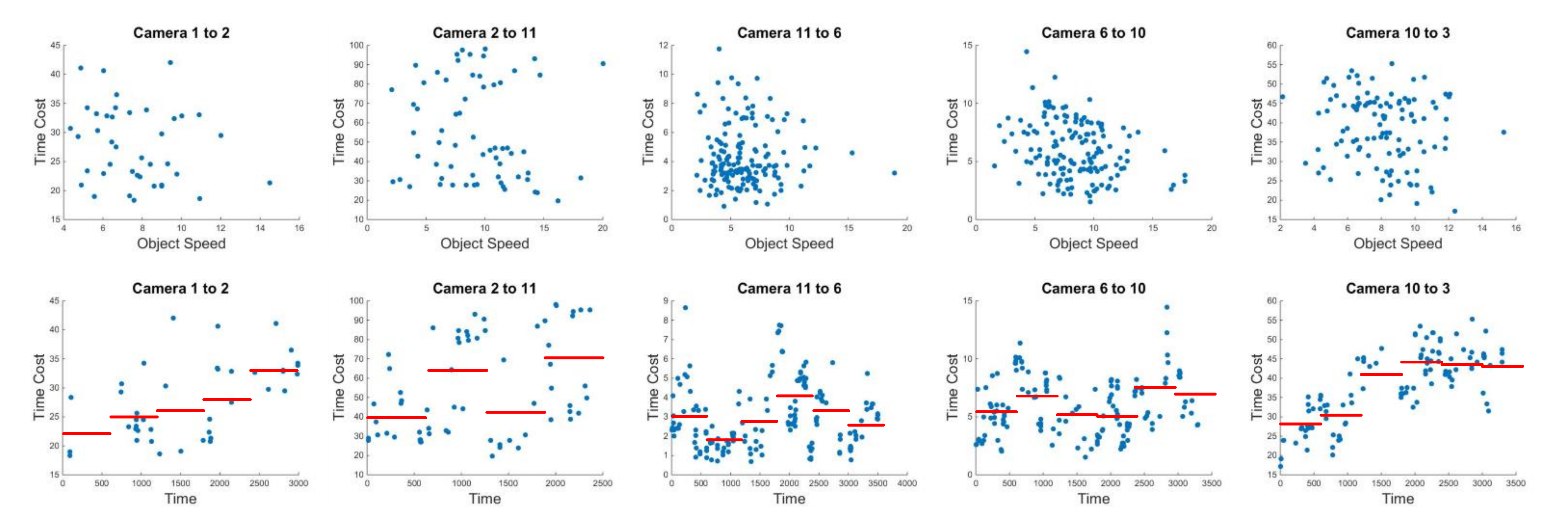}}
  \caption{Time cost statistics scatter plot.
  The top part are the time cost vs. object speed plots.
  The bottom part are the time cost vs. record time plots.
  The red lines in the bottom are the mean time cost in each 600-second time slot.}
  \label{fig:figure7}
\end{figure*}

However, according to the statistics on the our dataset as shown in Fig.~\ref{fig:figure7}, the above two model cannot be directly applied to our scenario.
We select 5 sequential edges in the surveillance network and plot the records in about one hour from 15:59:58 to 16:59:58.
In Fig.~\ref{fig:figure7}, the top row are the time cost vs. object speed plots.
We can find that the time costs are not linearly correlated with the speed of objects.
Because we can only obtain the speed at the cameras, yet cannot know the speed between the cameras.
The behaviors of vehicles between cameras are unpredictable with only surveillance videos.
The traffic lights, pedestrians, traffic jams make the actual model very complex.
So the linear model of time cost and speed would fail in our scenario.

The bottom part of Fig.~\ref{fig:figure7} illustrates the time cost vs. record time plots.
From the observation on this part, we find that in different time intervals the travel times of different vehicles change slightly.
In this case, we use a slot-mean model to build the weights. We segment the whole time line into time slots with the fixed length.
Supposing that set $C = \{c_{k,l}\}$ contains the time cost records on edge $e_{i,j}$ that fall in the time slot $k$.
We have the mean time cost $m_{i,j,k}$:
\begin{equation}
m_{i,j,k} = \frac{\sum^{|C|}_{l=1} c_{k,l} }{|C|}
\end{equation}

In each time slot $k$, $m_{i,j,k}$ is used as a parameter of the weight function.
In addition, we use $\tau_{i,j,k}$ as the other parameter of the weight which is computed as follow:
\begin{equation}
\tau_{i,j,k} = \sqrt{\frac{\sum^{|C|}_{l=1} (c_{k,l} - m_{i,j,k})^2 }{|C|}}
\end{equation}

After computing $(m_{i,j,k}, \tau_{i,j,k})$ on all time slots, we have a step function for the weight vector $w_{i,j}(x)=(m_{i,j}(x), \tau_{i,j}(x))^T$ on the edge:
\begin{equation}
w_{i,j}(x)=(m_{i,j}(x), \tau_{i,j}(x)) = \sum^{t}_{k=1}\chi_{i,j,k} (x)\cdot(m_{i,j,k} \ \tau_{i,j,k})
\end{equation}
where
\begin{equation}
\chi_{i,j,k} (x)=
\begin{cases}
1 & \text{if $x$ is in time slot $k$,}\\
0 & \text{else.}
\end{cases}
\end{equation}
where $x$ is an object metadata instance in the start camera $i$ of edge $e_{i,j}$, $t$ is the total number of time slots.
%In the search phase, given a metadata instance, we can use the weight $w_{i,j}(\cdot)$ to estimate its time cost on $e_{i,j}$.
All weight functions on the edges constitute the temporal weight set $W_t$ of graph $G$.

%Section 2.2
\section{Functional Modules}
\subsection{Vehicle Crawler}
\label{subsec:crawler}
The vehicle crawler aims to detect and crop vehicle images from video frames streamed by the surveillance network.
It plays a similar role to the conventional web crawler of the Internet search engines, which crawls and downloads webpages from  the World Wide Web.

To effectively locate the vehicles in the video frames, we adopt the state-of-the-art deep learning based object detection model, i.e., Faster R-CNN~\cite{ren2015faster}.
Faster R-CNN contains two Convolutional Neural Network (CNN) based parts.
The first is the Region Proposal Network, which is a Fully Convolutional Network (FCN) to generate object proposals from the input frames.
The second is a fully connected network to regress the bounding boxes of objects and the corresponding categories.
To achieve precise vehicle detection, we adopt a ResNet-50~\cite{he2016deep} based Faster R-CNN structure which is pretrained on the ImageNet dataset~\cite{russakovsky2015imagenet}.
Then, the network is finetuned on large-scale vehicle bounding boxes from surveillance videos annotated by ourselves.
After detection, a nearest neighbor tracking algorithm is adopted to associate vehicle bounding boxes of the same vehicle between neighbor frames.
In our implementation, the Faster R-CNN is deployed on the GPU servers to achieve efficient the vehicle detection.

For each track, it is assigned a unique vehicle ID under the corresponding camera.
The first frame of the track, the track length, and the sequence of pixel coordinates are recorded into the metadata, while the track that is shorter than 5 will be discarded.
After that, we use the off-the-shelf plate recognition tool to extract the plate numbers with a confidence measure.
If the tool cannot recognize the plate or return a very low confidence, the plate will be assigned as UNAVAL which means unavailable.
At last, the vehicle track metadata is appended to the camera table, meanwhile the image sequences of the track is stored on the vehicle storage server.

%Section 2.4
\subsection{Vehicle Indexer}
\label{subsec:index}
The vehicle indexer contains two functions: the first is multi-grained visual feature extraction, the second is multi-level index construction.

For the vehicle tracks, we extract the appearance based coarse representation and the license plate based fine-grained feature.
To learn discriminative and robust feature of vehicle appearance, we adopt the ResNet-50~\cite{he2016deep} pretrained on ImageNet~\cite{russakovsky2015imagenet} as the basic network.
The network is finetuned on the VeRi dataset~\cite{liu2016vehiclereid} with a multi task loss function, which contains a cross entropy loss and a contrastive loss~\cite{chopra2005learning}.
To learn effective plate feature, a ResNet-18 based siamese neural network for plate verification is trained on massive license plate pairs as in~\cite{liu2016provid} .
The above two feature extractor are deployed on the GPU servers for efficiency.
In the implementation, we use the 2048-D ``pool5'' layer of ResNet-50 and the 1024-D ``conv3'' layer of ResNet-18 as the appearance feature and plate feature, respectively.
For the images in the track, the features are extracted separately and fused by average pooling, which means that each vehicle track has a 2048-D coarse-grained feature and a 1024-D fine-grained feature.

After feature extraction, we build a two-level index for vehicle tracks with the state-of-the-art approximate nearest neighbor index algorithm, i.e., FLANN~\cite{muja2014flann}, due to its high efficiency.
The level-1 index is built on the appearance feature vectors, while the level-2 is built on the plate feature vectors.

%Section 2.4
\subsection{Vehicle Searcher}
\label{subsec:searcher}
In this section, we discuss the main procedures of online vehicle search.
Given a vehicle image cropped by a user and a time interval, a list of candidate target vehicles and their states will be returned, as shown in Figure~\ref{fig:figure1}.
As mention before, the progressive search contains two aspects:
%from-coarse-to-fine search with multi-grained features and from-near-to-distant search with spatiotemporal context.

\subsubsection{From-coarse-to-fine feature matching}
\label{subsubsec:match}
Vehicle search is generally an one-to-N feature matching problem, in which the similarity between the query and the gallery is estimated and ranked to find the most similar target vehicle to the query.
During searching, the query image or track is fed into the feature extraction module to extract its visual feature and plate feature as in Section~\ref{subsec:index}.
Then the visual feature of query is searched with the level-1 index to obtain the coarse similarity, $S_c$, between the query vehicle and the gallery vehicle.
Similarly, the fine similarity, $S_f$, is obtained with the level-2 index using the plate feature.
With the above two similarity scores, the visual similarity between the query vehicle, $V_q$, and one gallery vehicle, $V_g$ is:
\begin{equation}
S_v = \lambda \times S_c + (1-\lambda) \times S_f,
\end{equation}
where $\lambda$ is a hyper-parameter to balance the two scores.

In addition to the visual similarity, we also explore the spatiotemporal similarity between the query and the gallery.
Given the metadata of $V_q$, and $V_g$, we can obtain their spatial distance, $D_s$, and temporal distance, $D_t$ as
\begin{equation}
\begin{split}
&D_s = |L(c(V_q)) - L(c(V_g))|, \\
&D_t = |T(V_q) - T(V_g)|,
\end{split}
\end{equation}
where $c(\cdot)$ is the operation to get the camera ID of a vehicle, $L(\cdot)$ is the location of a camera, and $T(\cdot)$ .
Then, we adopt a two-layer fully connected neural network, i.e. the multi-layer perceptron (MLP), $F(\cdot)$,to model the spatiotemporal similarity fo $V_q$, and $V_g$.
The input and output dimensions of the two fully connected layers are (2, 64) and (64, 1), respectively.
The activation functions of the two layers are ReLU and Sigmoid, respectively.
The spatiotemporal similarity, $S_{st}$, is denoted as
\begin{equation}
S_{st} = F([D_s, D_t]),
\end{equation}
where $[\cdot, \cdot]$ is the concatenation of two elements.

At last, to effectively integrate the visual similarity, $S_v$, and spatiotemporal similarity $S_{st}$, we exploit a fully connected layer with sigmoid activation, $G(\cdot)$, to learn the suitable fusion parameter.
So, the final similarity can be computed by
\begin{equation}
S = G([S_v, S_{st}]).
\end{equation}
The neural networks $F(\cdot)$ and $G(\cdot)$ are trained with the binary cross entropy loss, which can guide the model to determine whether the query and one gallery are the same vehicle or not.
During searching, the results are ranked by the similarity scores $\{S_{q, g}\}$ between the query and the set of gallery vehicles.

\subsubsection{From-near-to-distant search}
\label{subsubsec:pro}
To achieve efficient vehicle search, we utilize the camera neighboring graph, $G$, to achieve the from-near-to-distant search.
Given the camera ID of the query, we traverse $G$ in the breadth-first manner.
It means that the query vehicle is matched first to the vehicles in the nearest neighboring cameras then to the distant ones.
After each traverse of current neighboring cameras, a list of candidate results is returned.
The results will update with the traverse of $G$ but the length of the list remains constant, which guarantees the most similar results can be shown to users.

\section{Experiments}
\label{sec:exp}

\subsection{Dataset}
\label{sec:data}
In this paper, we compare the proposed PVSS to different vehicle search methods on the VeRi dataset~\cite{Liu2018PROVID}.
The VeRi dataset is collected from 20 surveillance cameras in a real-world surveillance network, which contains about 50,000 images and 9000 tracks of 776 vehicles.
Each vehicle in the VeRi dataset is labeled with various attributes, such as 10 types of colors and 9 categories.
Moreover, the license number plates of vehicles are annotated for more precise vehicle search.
Furthermore, the context, such as the spatiotemporal information and the topology of the surveillance network, and distances are annotated.
Therefore, it is suitable to evaluate the proposed progressive vehicle search system.

\subsection{Experimental Settings}
As the similar settings in~\cite{Liu2018PROVID}, cross-camera matching is performed, which means that one vehicle image from one camera is used as the query to search for images of the same vehicle captured by other cameras.
Vehicle matching is in an track-to-track manner, which means units of the query set and the gallery are both tracks of vehicles cropped from surveillance videos.
In our experiments, we use 1,678 query tracks and 2,021 testing tracks as in~\cite{Liu2018PROVID}.

To evaluate the accuracy of the methods, the HIT@1 (precision at rank 1), and HIT@5 (precision at rank 5) are adopted.
In addition, since the query has more than one ground truth, the precision and recall should be considered in our experiments.
Hence, we also use mean average precision to evaluate the comprehensive performance as in~\cite{Liu2018PROVID}.
The average precision (AP) is computed for each query as
\begin{equation}\label{equ:equation4}
  AP = \frac{\sum_{k=1}^n P(k)\times gt(k)}{N_{gt}},
\end{equation}
where $n$ and $N_{gt}$ are the numbers of tests and ground truths respectively, $P(k)$ is the precision at the $k$-th position of the results, and $gt(k)$ is an indicator function that equals to $1$ if the $k$th result is correctly matched and $0$ otherwise.
Over all queries, the mean Average Precision (mAP) is formulated as
\begin{equation}\label{equ:equation5}
  mAP = \frac{\sum_{q=1}^Q AP(q)}{Q},
\end{equation}
in which $Q$ is the number of queries.

\subsection{Comparison with Vehicle Re-Id Methods}
In this section, we first compare the appearance based search component in PVSS with five appearance-based vehicle Re-Id methods.
Among them, methods 1) and 2) are two vehicle Re-Id methods, while methods 3) and 4) are two state-of-the-art approaches for video-based person Re-Id.
Then we compare the complete progressive vehicle search system with three state-of-the-art multi-modal data based approaches, which utilize visual features, plate features, and spatiotemporal data.
The details of all methods are as follows:

1) \textbf{Fusion of color and attribute (FACT)~\cite{liu2016vehiclereid}.}
This method is the baseline method on the VeRi dataset, which integrates hand-crafted features, e.g., SIFT and Color Name, with attributes extracted by GoogleNet.

2) \textbf{Progressive vehicle search (Progressive)~\cite{liu2016provid}.}
This is a progressive vehicle search framework, which uses appearance features and plate verification for vehicle matching and refines the results with spatiotemporal information.

3) \textbf{Identity feature with LSTM (ResNet + LSTM).}
This approach adopts the CNN+LSTM which is the state-of-the-art method for video-based person Re-Id~\cite{yan2016person}.
It can model dynamic patterns of persons like actions and gaits for person Re-Id.

4) \textbf{Top-push Distance Learning (TDL)~\cite{you2016top}.}
This method is one of the state-of-the-art metric learning methods for video-based person Re-Id.
We use the identity features extracted by ResNet as the basic features.
Then the TDL method is used to aggregate and map the original features into the latent space.

5) \textbf{Appearance-based search in PVSS (PVSS-App).}
This is a part of PVSS, which use only the appearance features for vehicle search.

6) \textbf{Orientation Invariant Feature Embedding and Spatial Temporal Regularization (OIFE + STR)~\cite{Wang_2017_ICCV}.}
This method proposes an Orientation Invariant Feature Embedding model to learn 20 landmarks and extract both local and global features from vehicle images.

7) \textbf{Siamese-CNN and Path-LSTM (SC + P-LSTM)-~\cite{shen2017learning}.}
This approach exploits two ResNets~\cite{he2016deep} in a siamese structure to learn visual feautres of vehicles and a one-layer LSTM to model the spatiotemporal context.

8) \textbf{PROgressive Vehicle re-ID (PROVID~\cite{Liu2018PROVID}).}
This progressive vehicle search framework search for vehicle in a three-step way: appearance-based coarse filtering, license plate-based fine search, and spatiotemporal re-ranking.

9) \textbf{PVSS-App-Plate.}
This is a part of the proposed PVSS, which use the appearance and plate features for vehicle search.

10) \textbf{PVSS.}
This is the complete progressive vehicle search system proposed in our paper.

\begin{table}[t]
\begin{center}
\caption{The results of vehicle Re-Id methods on VIVID dataset.}
\label{tab:table3}
\begin{tabular}{lccc}
    \hline
    methods                     								& mAP             		& HIT@1             	&  HIT@5 \\
    \hline
    FACT~\cite{liu2016vehiclereid}  						& 18.00           &  52.44            &  72.29 \\
    Progressive~\cite{liu2016provid} 						& 25.11           &  61.26            &  75.98 \\
    ResNet + LSTM~\cite{yan2016person}					& 28.11	          &  56.20	          &  79.14 \\
    TDL~\cite{you2016top}        			          			& 35.65	          &  69.61	          &  88.02 \\
    PVSS-App        			                  				& \textbf{51.00}           &  \textbf{85.64}            &  \textbf{95.35} \\
    \hline
    \hline
    OIFE + STR~\cite{Wang_2017_ICCV}					& 51.42	         	&  68.30	          	&  89.07 \\
    SC + Path-LSTM~\cite{shen2017learning}         			& 58.27           		&  83.49            	&  90.04 \\
    PROVID~\cite{Liu2018PROVID}            				& 53.42           		&  81.56            	&  95.11 \\
    PVSS-App-Plate           								& 61.12           		&  89.69            	&  96.31 \\
    PVSS                         								& \textbf{62.62}         & \textbf{90.58}     	& \textbf{97.14}  \\
\hline
\end{tabular}
\end{center}%\vspace{-5mm}
\end{table}

\tabcolsep 12pt
%\cmidrule(l){2-4}%
\renewcommand\arraystretch{1.3}
\begin{center}
{\footnotesize{\bf Table 2.} The results of vehicle Re-Id methods on VIVID.}\\
\vspace{2mm}
\footnotesize{
\begin{tabular}{lccc}
    \hline
    methods                     								& mAP             		& HIT@1             	&  HIT@5 \\
    \hline
    FACT~\cite{liu2016vehiclereid}  						& 18.00           &  52.44            &  72.29 \\
    Progressive~\cite{liu2016provid} 						& 25.11           &  61.26            &  75.98 \\
    ResNet + LSTM~\cite{yan2016person}					& 28.11	          &  56.20	          &  79.14 \\
    TDL~\cite{you2016top}        			          			& 35.65	          &  69.61	          &  88.02 \\
    PVSS-App        			                  				& \textbf{51.00}           &  \textbf{85.64}            &  \textbf{95.35} \\
    \hline
    \hline
    OIFE + STR~\cite{Wang_2017_ICCV}					& 51.42	         	&  68.30	          	&  89.07 \\
    SC + Path-LSTM~\cite{shen2017learning}         			& 58.27           		&  83.49            	&  90.04 \\
    PROVID~\cite{Liu2018PROVID}            				& 53.42           		&  81.56            	&  95.11 \\
    PVSS-App-Plate           								& 61.12           		&  89.69            	&  96.31 \\
    PVSS                         								& \textbf{62.62}         & \textbf{90.58}     	& \textbf{97.14}  \\
\hline
\end{tabular}
}
\end{center}
Table 2 lists the mAP, HIT@1, and HIT@5 of approaches.
For appearance-only methods, we can find that the traditional methods, i.e., FACT and Progressive, are worse than deep learning based methods.
This is because the hand-crafted features cannot effectively model the appearance of a vehicle and comprehensively represent the vehicles.
By comparing LSTM-based methods with other deep learning-based models, we can see that LSTM-based methods obtain worse results.
Although LSTM can model dynamic representation from action or gait for video-based person Re-Id, it may be failed for video-based vehicle Re-Id.
The TDL performs better than the LSTM-based method, while our appearance-based part in PVSS-App achieves the best results.
For the multi-modal methods, the OIFE + STR and SC + Path-LSTM obtain worse results than the proposed PVSS-App-Plate, because these two methods neglect the license plates to uniquely identify vehicles.
Moreover, by incorporating spatiotemporal context, the PVSS outperforms other multi-modal search methods and achieves the best results.

\section{Conclusions}
\label{sec:con}
This paper proposes PVSS, a progressive vehicle search system which can crawl and index vehicles captured by large-scale surveillance networks and provide vehicle search services for users.
For the vehicle crawler, the vehicle detection and tracking algorithms are adopted to crop vehicle images from surveillance videos.
Then, vehicle images are fed into the vehicle indexer to extract multi-grained visual features, which are utilized to build a multi-level index for vehicle search.
In the online search stage, the target vehicle is searched in a from-coarse-to-fine manner with the multi-level index and in a from-near-to-distant way based on the spatiotemporal context of the surveillance network.
Extensive evaluations on the VeRi dataset show the excellent performance of the PVSS.

\section*{Acknowledgment}
This work was supported in part by the Funds for International Cooperation and Exchange of the National Natural Science Foundation of China
(No. 61720106007),
the NSFC-Guangdong Joint Fund (No. U1501254),
the National Key Research and Development Plan (No. 2016YFC0801005),
%the Funds for Creative Research Groups of China (No. 61421061),
the National Natural Science Foundation of China (No. 61602049),
and the 111 Project (B18008).

\bibliographystyle{unsrt}
\small
\bibliography{JCST_submit}

\label{last-page}
\end{multicols}
\label{last-page}
\end{document}